# Probabilities of Causation: Bounds and Identification

Jin Tian and Judea Pearl
Cognitive Systems Laboratory
Computer Science Department
University of California, Los Angeles, CA 90024
{jtian, judea }@cs.ucla.edu

## Abstract

This paper deals with the problem of estimating the probability that one event was a cause of another in a given scenario. Using structural-semantical definitions of the probabilities of necessary or sufficient causation (or both), we show how to optimally bound these quantities from data obtained in experimental and observational studies, making minimal assumptions concerning the data-generating process. In particular, we strengthen the results of Pearl (1999) by weakening the data-generation assumptions and deriving theoretically sharp bounds on the probabilities of causation. These results delineate precisely how empirical data can be used both in settling questions of attribution and in solving attribution-related problems of decision making.

## 1 Introduction

Assessing the likelihood that one event *was the cause* of another guides much of what we understand about (and how we act in) the world. For example, few of us would take aspirin to combat headache if it were not for our conviction that, with high probability, it was aspirin that "actually caused" relief in previous headache episodes. [Pearl, 1999] gave counterfactual definitions for the probabilities of *necessary* or *sufficient* causation (or both) based on structural model semantics, which defines counterfactuals as quantities derived from modifiable sets of functions [Galles and Pearl, 1997, Galles and Pearl, 1998, Halpern, 1998, Pearl, 2000, chapter 7].

The central aim of this paper is to estimate probabilities of causation from frequency data, as obtained in experimental and observational statistical studies. In general, such probabilities are *non-identifiable*, that is, non-estimable from frequency data alone. One factor that hinders identifiability is confounding – the cause and the effect may both be influenced by a third factor. Moreover, even in the absence of confounding, probabilities of causation are sensitive to the data-generating process, namely, the functional relationships that connect causes and effects [Robins and Greenland, 1989, Balke and Pearl, 1994]. Nonetheless, useful information in the form of *bounds* on the probabilities of causation can be extracted from empirical data without actually knowing the data-generating process. We show that these bounds improve when data from observational and experimental studies are combined. Additionally, under certain assumptions about the data-generating process (such as exogeneity and monotonicity), the bounds may collapse to point estimates, which means that the probabilities of causation are *identifiable* – they can be expressed in terms of probabilities of observed quantities. These estimates often appear in the literature as measures of *attribution*, and our analysis thus explicates the assumptions that must be ascertained before those measures can legitimately be interpreted as probabilities of causation.

The analysis of this paper extends the results reported in [Pearl, 1999] [Pearl, 2000, pp. 283-308]. Pearl derived bounds and identification conditions under certain assumptions of exogeneity and monotonicity, and this paper narrows his bounds and weakens his assumptions. In particular, we show that for most of Pearl's results, the assumption of strong exogeneity can be replaced by weak exogeneity (to be defined in Section 3.3). Additionally, we show that the point estimates that Pearl obtained under the assumption of monotonicity (Definition 6) constitute valid lower bounds when monotonicity is not assumed. Finally, we prove that the bounds derived by Pearl, as well as those provided in this paper are *sharp*, that is, they cannot be improved without strengthening the assumptions. We illustrate the use of our results in the context of legal disputes (Section 4) and personal decision making (Section 5).

## 2 Probabilities of Causation: Definitions

In this section, we present the definitions for the three aspects of causation as defined in [Pearl, 1999]. We use the language of counterfactuals in its structural model semantics, as given in Balke and Pearl (1995), Galles and Pearl (1997, 1998), and Halpern (1998). We use $Y_x = y$ to denote the counterfactual sentence "Variable $Y$ would have the value $y$, had $X$ been $x$." The structural model interpretation of this sentence reads: "Deleting the equation for $X$ from the model and setting the value of $X$ to a constant $x$ will yield a solution in which variable $Y$ will take on the value $y$."

One property that the counterfactual relationships satisfy is the *consistency condition* [Robins, 1987]:

$$(X = x) \Rightarrow (Y_x = Y) \quad (1)$$

stating that if we intervene and set the experimental conditions $X = x$ equal to those prevailing before the intervention, we should not expect any change in the response variable $Y$. This property will be used in several derivations of this section and Section 3. For detailed exposition of the structural account and its applications see [Pearl, 2000, chapter 7]. For notational simplicity, we limit the discussion to binary variables; extension to multi-valued variables are straightforward (see Pearl 2000, p. 286, footnote 5).

**Definition 1** (*Probability of necessity (PN)*)
*Let $X$ and $Y$ be two binary variables in a causal model $M$, let $x$ and $y$ stand for the propositions $X = true$ and $Y = true$, respectively, and $x'$ and $y'$ for their complements. The probability of necessity is defined as the expression*

$$\begin{aligned} PN &\triangleq P(Y_{x'} = false \mid X = true, Y = true) \\ &\triangleq P(y'_{x'}|x,y) \end{aligned} \quad (2)$$

In other words, PN stands for the probability that event $y$ would not have occurred in the absence of event $x$, $y'_{x'}$, given that $x$ and $y$ did in fact occur.

Note that lower case letters (e.g., $x, y$) stand for propositions (or events). Note also the abbreviations $y_x$ for $Y_x = true$ and $y'_x$ for $Y_x = false$. Readers accustomed to writing "$A > B$" for the counterfactual "$B$ if it were $A$" can translate Eq. (2) to read $PN \triangleq P(x' > y'|x,y)$.

PN has applications in epidemiology, legal reasoning, and artificial intelligence (AI). Epidemiologists have long been concerned with estimating the probability that a certain case of disease is *attributable* to a particular exposure, which is normally interpreted counterfactually as "the probability that disease would not have occurred in the absence of exposure, given that disease and exposure did in fact occur." This counterfactual notion is also used frequently in lawsuits, where legal responsibility is at the center of contention (see Section 4).

**Definition 2** (*Probability of sufficiency (PS)*)

$$PS \triangleq P(y_x|y', x') \quad (3)$$

PS finds applications in policy analysis, AI, and psychology. A policy maker may well be interested in the dangers that a certain exposure may present to the healthy population [Khoury et al., 1989]. Counterfactually, this notion is expressed as the "probability that a healthy unexposed individual would have gotten the disease had he/she been exposed." In psychology, PS serves as the basis for Cheng's (1997) causal power theory [Glymour, 1998], which attempts to explain how humans judge causal strength among events. In AI, PS plays a major role in the generation of explanations [Pearl, 2000, pp. 221-223].

**Definition 3** (*Probability of necessity and sufficiency (PNS)*)

$$PNS \triangleq P(y_x, y'_{x'}) \quad (4)$$

PNS stands for the probability that $y$ would respond to $x$ both ways, and therefore measures both the sufficiency and necessity of $x$ to produce $y$.

Although none of these quantities is sufficient for determining the others, they are not entirely independent, as shown in the following lemma.

**Lemma 1** *The probabilities of causation satisfy the following relationship* [Pearl, 1999] :

$$PNS = P(x,y)PN + P(x',y')PS \quad (5)$$

Since all the causal measures defined above invoke conditionalization on $y$, and since $y$ is presumed affected by $x$, the antecedent of the counterfactual $y_x$, we know that none of these quantities is identifiable from knowledge of frequency data alone, even under condition of no confounding. However, useful information in the form of bounds may be derived for these quantities from frequency data, especially when knowledge about *causal effects* $P(y_x)$ and $P(y_{x'})$ is also available[1]. Moreover, under some general assumptions about the data-generating process, these quantities may even be identified.

## 3 Bounds and Conditions of Identification

In this section we will assume that experimental data will be summarized in the form of the causal effects $P(y_x)$ and $P(y_{x'})$, and nonexperimental data will be summarized in the form of the joint probability function: $P_{XY} = \{P(x,y), P(x',y), P(x,y'), P(x',y')\}$.

### 3.1 Linear programming formulation

Since every causal model induces a joint probability distribution on the four binary variables: $X$, $Y$, $Y_x$ and $Y_{x'}$, specifying the sixteen parameters of this distribution would suffice for computing the PN, PS, and PNS. Moreover, since $Y$ is a deterministic function of the other three variables, the problem is fully specified by the following set of eight parameters:

$$
\begin{aligned}
p_{111} &= P(y_x, y_{x'}, x) &= P(x, y, y_{x'}) \\
p_{110} &= P(y_x, y_{x'}, x') &= P(x', y, y_x) \\
p_{101} &= P(y_x, y'_{x'}, x) &= P(x, y, y'_{x'}) \\
p_{100} &= P(y_x, y'_{x'}, x') &= P(x', y', y_x) \\
p_{011} &= P(y'_x, y_{x'}, x) &= P(x, y', y_{x'}) \\
p_{010} &= P(y'_x, y_{x'}, x') &= P(x', y, y'_x) \\
p_{001} &= P(y'_x, y'_{x'}, x) &= P(x, y', y'_{x'}) \\
p_{000} &= P(y'_x, y'_{x'}, x') &= P(x', y', y'_x)
\end{aligned}
$$

where we have used the consistency condition Eq. (1). These parameters are further constrained by the probabilistic equality

$$\sum_{i=0}^{1}\sum_{j=0}^{1}\sum_{k=0}^{1} p_{ijk} = 1$$

$$p_{ijk} \geq 0 \text{ for } i,j,k \in \{0,1\} \quad (6)$$

In addition, the nonexperimental probabilities $P_{XY}$ impose the constraints:

$$
\begin{aligned}
p_{111} + p_{101} &= P(x,y) \\
p_{011} + p_{001} &= P(x,y') \quad (7) \\
p_{110} + p_{010} &= P(x',y)
\end{aligned}
$$

---
[1] The causal effects $P(y_x)$ and $P(y_{x'})$ can be estimated reliably from controlled experimental studies, and from certain observational (i.e., nonexperimental) studies which permit the control of confounding through adjustment of covariates [Pearl, 1995].

and the causal effects, $P(y_x)$ and $P(y_{x'})$, impose the constraints:

$$
\begin{aligned}
P(y_x) &= p_{111} + p_{110} + p_{101} + p_{100} \\
P(y_{x'}) &= p_{111} + p_{110} + p_{011} + p_{010}
\end{aligned} \quad (8)
$$

The quantities we wish to bound are:

$$
\begin{aligned}
PNS &= p_{101} + p_{100} & (9) \\
PN &= p_{101}/P(x,y) & (10) \\
PS &= p_{100}/P(x',y') & (11)
\end{aligned}
$$

Optimizing the functions in (9)–(11), subject to equality constraints, defines a linear programming (LP) problem that lends itself to closed-form solution. Balke (1995, Appendix B) describes a computer program that takes symbolic descriptions of LP problems and returns symbolic expressions for the desired bounds. The program works by systematically enumerating the vertices of the constraint polygon of the dual problem. The bounds reported in this paper were produced (or tested) using Balke's program, and will be stated here without proofs; their correctness can be verified by manually enumerating the vertices as described in [Balke, 1995, Appendix B]. These bounds are guaranteed to be sharp because the optimization is global.

### 3.2 Bounds with no assumptions

#### 3.2.1 Given nonexperimental data

Given $P_{XY}$, constraints (6) and (7) induce the following upper bound on PNS:

$$0 \leq PNS \leq P(x,y) + P(x',y'). \quad (12)$$

However, PN and PS are not constrained by $P_{XY}$.

These constraints also induce bounds on the causal effects $P(y_x)$ and $P(y_{x'})$:

$$
\begin{aligned}
P(x,y) &\leq P(y_x) \leq 1 - P(x,y') \\
P(x',y) &\leq P(y_{x'}) \leq 1 - P(x',y')
\end{aligned} \quad (13)
$$

#### 3.2.2 Given causal effects

Given constraints (6) and (8), the bounds induced on PNS are:

$$\max[0, P(y_x) - P(y_{x'})] \leq PNS \leq \min[P(y_x), P(y'_{x'})] \quad (14)$$

with no constraints on PN and PS.

#### 3.2.3 Given both nonexperimental data and causal effects

Given the constraints (6), (7) and (8), the following bounds are induced on the three probabilities of cau-

sation:

$$\max \left\{ \begin{array}{c} 0 \\ P(y_x) - P(y_{x'}) \\ P(y) - P(y_{x'}) \\ P(y_x) - P(y) \end{array} \right\} \leq PNS \qquad (15)$$

$$PNS \leq \min \left\{ \begin{array}{c} P(y_x) \\ P(y'_{x'}) \\ P(x,y) + P(x',y') \\ P(y_x) - P(y_{x'}) + P(x,y') + P(x',y) \end{array} \right\} \qquad (16)$$

$$\max \left\{ \begin{array}{c} 0 \\ \frac{P(y) - P(y_{x'})}{P(x,y)} \end{array} \right\} \leq PN \leq \min \left\{ \begin{array}{c} 1 \\ \frac{P(y'_{x'}) - P(x',y')}{P(x,y)} \end{array} \right\} \qquad (17)$$

$$\max \left\{ \begin{array}{c} 0 \\ \frac{P(y_x) - P(y)}{P(x',y')} \end{array} \right\} \leq PS \leq \min \left\{ \begin{array}{c} 1 \\ \frac{P(y_x) - P(x,y)}{P(x',y')} \end{array} \right\} \qquad (18)$$

Thus we see that some information about PN and PS can be extracted without making any assumptions about the data-generating process. Furthermore, combined data from both experimental and nonexperimental studies yield information that neither study alone can provide.

### 3.3 Bounds under exogeneity (no confounding)

**Definition 4** (*Exogeneity*)
*A variable $X$ is said to be exogenous for $Y$ in model $M$ iff*

$$P(y_x) = P(y|x) \quad \text{and} \quad P(y_{x'}) = P(y|x'). \qquad (19)$$

*In words, the way $Y$ would potentially respond to experimental conditions $x$ or $x'$ is independent of the actual value of $X$.*

Eq. (19) is also known as "no-confounding" [Robins and Greenland, 1989], "as if randomized," or "weak ignorability" [Rosenbaum and Rubin, 1983].

Combining Eq. (19) with the constraints of (6)-(8), the linear programming optimization (Section 3.1) yields the following results:

**Theorem 1** *Under condition of exogeneity, the three probabilities of causation are bounded as follows:*

$$\max[0, P(y|x) - P(y|x')] \leq PNS \leq \min[P(y|x), P(y'|x')] \qquad (20)$$

$$\frac{\max[0, P(y|x) - P(y|x')]}{P(y|x)} \leq PN \leq \frac{\min[P(y|x), P(y'|x')]}{P(y|x)} \qquad (21)$$

$$\frac{\max[0, P(y|x) - P(y|x')]}{P(y'|x')} \leq PS \leq \frac{\min[P(y|x), P(y'|x')]}{P(y'|x')} \qquad (22)$$

[Pearl, 1999] derived Eqs. (20)-(22) under a stronger condition of exogeneity (see Definition 5). We see that under the condition of no-confounding the lower bound for PN can be expressed as

$$PN \geq 1 - \frac{1}{P(y|x)/P(y|x')} \triangleq 1 - \frac{1}{RR} \qquad (23)$$

where $RR \triangleq P(y|x)/P(y|x')$ is called *relative risk* in epidemiology. Courts have often used the condition $RR > 2$ as a criterion for legal responsibility [Bailey et al., 1994]. Eq. (23) shows that this practice represents a conservative interpretation of the "more probable than not" standard (assuming no confounding); PN must indeed be higher than 0.5 if RR exceeds 2.

#### 3.3.1 Bounds under strong exogeneity

The condition of exogeneity, as defined in Eq. (19) is testable by comparing experimental and nonexperimental data. A stronger version of exogeneity can be defined as the joint independence $\{Y_x, Y_{x'}\} \perp\!\!\!\perp X$ which was called "strong ignorability" by Rosenbaum and Rubin (1983). Though untestable, such joint independence is implied when we assert the absence of factors that simultaneously affect exposure and outcome.

**Definition 5** (*Strong Exogeneity*)
*A variable $X$ is said to be strongly exogenous for $Y$ in model $M$ iff $\{Y_x, Y_{x'}\} \perp\!\!\!\perp X$, that is,*

$$\begin{array}{rcl} P(y_x, y_{x'}|x) & = & P(y_x, y_{x'}) \\ P(y_x, y'_{x'}|x) & = & P(y_x, y'_{x'}) \\ P(y'_x, y_{x'}|x) & = & P(y'_x, y_{x'}) \\ P(y'_x, y'_{x'}|x) & = & P(y'_x, y'_{x'}) \end{array} \qquad (24)$$

Remarkably, the added constraints introduced by strong exogeneity do not alter the bounds of Eqs. (20)-(22). They do, however, strengthen Lemma 1:

**Theorem 2** *If strong exogeneity holds, the probabilities PN, PS, and PNS are constrained by the bounds of Eqs. (20)-(22), and, moreover, PN, PS, and PNS are related to each other as follows* [Pearl, 1999]:

$$PN = \frac{PNS}{P(y|x)} \qquad (25)$$

$$PS = \frac{PNS}{P(y'|x')} \qquad (26)$$

### 3.4 Identifiability under monotonicity

**Definition 6** (*Monotonicity*)
*A variable $Y$ is said to be monotonic relative to variable $X$ in a causal model $M$ iff*

$$y'_x \wedge y_{x'} = \text{false} \qquad (27)$$

Monotonicity expresses the assumption that a change from $X =$ false to $X =$ true cannot, under any circumstance make $Y$ change from *true* to *false*. In epidemiology, this assumption is often expressed as "no prevention," that is, no individual in the population can be helped by exposure to the risk factor.

In the linear programming formulation of Section 3.1, monotonicity narrows the feasible space to the manifold:

$$p_{011} = 0$$
$$p_{010} = 0 \quad (28)$$

### 3.4.1 Given nonexperimental data

Under the constraints (6), (7), and (28), we find the same bounds for PNS as the ones obtained under no assumptions (Eq. (12)). Moreover, there are still no constraints on PN and PS. Thus, with nonexperimental data alone, the monotonicity assumption does not provide new information.

However, the monotonicity assumption induces sharper bounds on the causal effects $P(y_x)$ and $P(y_{x'})$:

$$P(y) \leq P(y_x) \leq 1 - P(x,y')$$
$$P(x',y) \leq P(y_{x'}) \leq P(y) \quad (29)$$

Compared with Eq. (13), the lower bound for $P(y_x)$ and the upper bound for $P(y_{x'})$ are tightened. The importance of Eq. (29) lies in providing a simple necessary test for the commonly made assumption of "no-prevention." These inequalities are sharp, in the sense that every combination of experimental and non-experimental data that satisfy these inequalities can be generated from some causal model in which $Y$ is monotonic in $X$. Alternatively, if the no-prevention assumption is theoretically unassailable, the inequalities of Eq. (29) can be used for testing the compatibility of the experimental and non-experimental data, namely, whether subjects used in clinical trials were sampled from the same target population, characterized by the joint distribution $P_{XY}$.

### 3.4.2 Given causal effects

Constraints (6), (8), and (28) induce no constraints on PN and PS, while the value of PNS is fully determined:

$$PNS = P(y_x, y'_{x'}) = P(y_x) - P(y_{x'})$$

That is, under the assumption of monotonicity, PNS can be determined by experimental data alone, although the joint event $y_x \wedge y'_{x'}$ can never be observed.

### 3.4.3 Given both nonexperimental data and causal effects

Under the constraints (6)–(8) and (28), the values of PN, PS, and PNS are all determined precisely.

**Theorem 3** *If $Y$ is monotonic relative to $X$, then PNS, PN, and PS are given by*

$$PNS = P(y_x, y'_{x'}) = P(y_x) - P(y_{x'}) \quad (30)$$
$$PN = P(y'_{x'}|x,y) = \frac{P(y) - P(y_{x'})}{P(x,y)} \quad (31)$$
$$PS = P(y_x|x',y') = \frac{P(y_x) - P(y)}{P(x',y')} \quad (32)$$

Eqs. (30)–(32) are applicable to situations where, in addition to observational probabilities, we also have information about the causal effects $P(y_x)$ and $P(y_{x'})$. Such information may be obtained either directly, through separate experimental studies, or indirectly, from observational studies in which certain identifying assumptions are deemed plausible (e.g., assumptions that permits identification through adjustment of covariates) [Pearl, 1995].

### 3.5 Identifiability under monotonicity and exogeneity

Under the assumption of monotonicity, if we further assume exogeneity, then $P(y_x)$ and $P(y_{x'})$ are identified through Eq. (19), and from theorem 3 we conclude that PNS, PN, and PS are all identifiable.

**Theorem 4** (*Identifiability under exogeneity and monotonicity*)
*If $X$ is exogenous and $Y$ is monotonic relative to $X$, then the probabilities PN, PS, and PNS are all identifiable, and are given by*

$$PNS = P(y|x) - P(y|x') \quad (33)$$
$$PN = \frac{P(y) - P(y|x')}{P(x,y)} = \frac{P(y|x) - P(y|x')}{P(y|x)} \quad (34)$$
$$PS = \frac{P(y|x) - P(y)}{P(x',y')} = \frac{P(y|x) - P(y|x')}{P(y'|x')} \quad (35)$$

These expressions are to be recognized as familiar measures of attribution that often appear in the literature. The r.h.s. of (33) is called "risk-difference" in epidemiology, and is also misnamed "attributable risk" [Hennekens and Buring, 1987, p. 87]. The probability of necessity, PN, is given by the *excess-risk-ratio* (ERR)

$$PN = \frac{P(y|x) - P(y|x')}{P(y|x)} = 1 - \frac{1}{RR} \quad (36)$$

often misnamed as the *attributable fraction, attributable-rate percent, attributed fraction for the exposed* [Kelsey et al., 1996, p. 38], or *attributable proportion* [Cole, 1997]. The reason we consider these labels to be misnomers is that ERR invokes purely statistical relationships, hence it cannot in itself serve to measure attribution, unless fortified with some causal assumptions. Exogeneity and monotonicity are the causal assumptions that endow ERR with attributional interpretation, and these assumptions are rarely made explicit in the literature on attribution.

The expression for PS is likewise quite revealing

$$PS = [P(y|x) - P(y|x')]/[1 - P(y|x')], \qquad (37)$$

as it coincides with what epidemiologists call the "relative difference" [Shep, 1958], which is used to measure the *susceptibility* of a population to a risk factor $x$. It also coincides with what Cheng calls "causal power" (1997), namely, the effect of $x$ on $y$ after suppressing "all other causes of $y$." See Pearl (1999) for additional discussions of these expressions.

To appreciate the difference between Eqs. (31) and (36) we can rewrite Eq. (31) as

$$\begin{aligned} PN &= \frac{P(y|x)P(x) + P(y|x')P(x') - P(y_{x'})}{P(y|x)P(x)} \\ &= \frac{P(y|x) - P(y|x')}{P(y|x)} + \frac{P(y|x') - P(y_{x'})}{P(x,y)} \end{aligned} \qquad (38)$$

The first term on the r.h.s. of (38) is the familiar ERR as in (36), and represents the value of PN under exogeneity. The second term represents the correction needed to account for $X$'s non-exogeneity, i.e. $P(y_{x'}) \neq P(y|x')$. We will call the r.h.s. of (38) by corrected excess-risk-ratio (CERR).

From Eqs. (33)–(35) we see that the three notions of causation satisfy the simple relationships given by Eqs. (25) and (26) which we obtained under the strong exogeneity condition. In fact, we have the following theorem.

**Theorem 5** *Monotonicity (27) and exogeneity (19) together imply strong exogeneity (24).*

### 3.6 Summary of results

Table 1 lists the best estimate of PN under various assumptions and various types of data—the stronger the assumptions, the more informative the estimates. We see that the excess-risk-ratio (ERR), which epidemiologists commonly identify with the probability of causation, is a valid measure of PN only when two assumptions can be ascertained: exogeneity (i.e., no confounding) and monotonicity (i.e., no prevention). When monotonicity does not hold, ERR provides merely a lower bound for PN, as shown in Eq. (21). (The upper bound is usually unity.) In the presence of confounding, ERR must be corrected by the additive term $[P(y|x') - P(y_{x'})]/P(x,y)$, as stated in (38). In other words, when confounding bias (of the causal effect) is positive, PN is higher than ERR by the amount of this additive term. Clearly, owing to the division by $P(x,y)$, the PN bias can be many times higher than the causal effect bias $P(y|x') - P(y_{x'})$. However, confounding results only from association between exposure and other factors that affect the outcome; one need not be concerned with associations between such factors and susceptibility to exposure, as is often assumed in the literature [Khoury et al., 1989, Glymour, 1998].

Table 1: PN as a function of assumptions (exogeneity or monotonicity) and available data (experimental or nonexperimental or both). ERR stands of the excess-risk-ratio and CERR is given in Eq. (38). The non-entries (—) represent vacuous bounds, that is, $0 \leq PN \leq 1$.

| Assumptions | | Data Available | | |
|---|---|---|---|---|
| Exo. | Mono. | Exp. | Non-exp. | Combined |
| + | + | ERR | ERR | ERR |
| + | – | bounds | bounds | bounds |
| – | + | — | — | CERR |
| – | – | — | — | bounds |

The last two rows in Table 1 correspond to no assumptions about exogeneity, and they yield vacuous bounds for PN when data come from either experimental or observational study. In contrast, informative bounds (17) or point estimates (38) are obtained when data from experimental and observational studies are combined. Concrete use of such combination will be illustrated in Section 4.

## 4 Example 1: Legal Responsibility

A lawsuit is filed against the manufacturer of drug $x$, charging that the drug is likely to have caused the death of Mr. A, who took the drug to relieve symptom $S$ associated with disease $D$.

The manufacturer claims that experimental data on patients with symptom $S$ show conclusively that drug $x$ may cause only minor increase in death rates. The plaintiff argues, however, that the experimental study is of little relevance to this case, because it represents the effect of the drug on *all* patients, not on patients like Mr. A who actually died while using drug $x$. Moreover, argues the plaintiff, Mr. A is unique in that he used the drug on his own voli-tion, unlike subjects in the experimental study who took the drug to comply with experimental protocols. To support this argument, the plaintiff furnishes non-experimental data indicating that most patients who chose drug $x$ would have been alive if it were not for the drug. The manufacturer counter-argues by stating that: (1) counterfactual speculations regarding whether patients would or would not have died are purely metaphysical and should be avoided, and (2) nonexperimental data should be dismissed a priori, on the ground that such data may be highly biased; for example, incurable terminal patients might be more inclined to use drug $x$ if it provides them greater symptomatic relief. The court must now decide, based on both the experimental and non-experimental studies, what the probability is that drug $x$ was in fact the cause of Mr. A's death.

Table 2: Frequency data (hypothetical) obtained in experimental and nonexperimental studies, comparing deaths (in thousands) among drug users ($x$) and non-users ($x'$).

|  | Experimental | | Nonexperimental | |
|---|---|---|---|---|
|  | $x$ | $x'$ | $x$ | $x'$ |
| Deaths($y$) | 16 | 14 | 2 | 28 |
| Survivals($y'$) | 984 | 986 | 998 | 972 |

The (hypothetical) data associated with the two studies are shown in Table 2. The experimental data provide the estimates

$$P(y_x) \quad = 16/1000 \quad = 0.016$$
$$P(y_{x'}) \quad = 14/1000 \quad = 0.014$$
$$P(y'_{x'}) \quad = 1 - P(y_{x'}) \quad = 0.986$$

The non-experimental data provide the estimates

$$P(y) \quad = 30/2000 \quad = 0.015$$
$$P(x,y) \quad = 2/2000 \quad = 0.001$$
$$P(x',y') \quad = 972/2000 \quad = 0.486$$

Since both the experimental and nonexperimental data are available, we can obtain bounds on all three probabilities of causation through Eqs. (15)–(18) without making any assumptions about the underlying mechanisms. The data in Table 2 imply the following numerical results:

$$0.002 \quad \leq PNS \leq \quad 0.016 \qquad (39)$$
$$1.0 \quad \leq PN \leq \quad 1.0 \qquad (40)$$
$$0.002 \quad \leq PS \leq \quad 0.031 \qquad (41)$$

These figures show that although surviving patients who didn't take drug $x$ have only less than 3.1% chance to die had they taken the drug, there is 100% assurance (barring sample errors) that those who took the drug and died would have survived had they not taken the drug. Thus the plaintiff was correct; drug $x$ was in fact responsible for the death of Mr. A.

If we assume that drug $x$ can only cause, but never prevent, death, Theorem 3 is applicable and Eqs. (30)–(32) yield

$$PNS = 0.002 \qquad (42)$$
$$PN = 1.0 \qquad (43)$$
$$PS = 0.002 \qquad (44)$$

Thus, we conclude that drug $x$ was responsible for the death of Mr. A, with or without the no-prevention assumption.

Note that a straightforward use of the experimental excess-risk-ratio would yield a much lower (and incorrect) result:

$$\frac{P(y_x) - P(y_{x'})}{P(y_x)} = \frac{0.016 - 0.014}{0.016} = 0.125 \qquad (45)$$

Evidently, what the experimental study does not reveal is that, given a choice, terminal patients stay away from drug $x$. Indeed, if there were any terminal patients who would choose $x$ (given the choice), then the control group ($x'$) would have included some such patients (due to randomization) and so the proportion of deaths among the control group $P(y_{x'})$ would have been higher than $P(x',y)$, the population proportion of terminal patients avoiding $x$. However, the equality $P(y_{x'}) = P(y,x')$ tells us that no such patients were present in the control group, hence (by randomization) no such patients exist in the population at large and therefore none of the patients who freely chose drug $x$ was a terminal case; all were susceptible to $x$.

The numbers in Table 2 were obviously contrived to show the usefulness of the bounds in Eqs. (15)-(18). Nevertheless, it is instructive to note that a combination of experimental and non-experimental studies may unravel what experimental studies alone will not reveal.

## 5 Example 2: Personal Decision Making

Consider the case of Mr. B, who is one of the surviving patients in the observational study of Table 2. Mr. B wonders how safe it would be for him to take drug $x$, given that he has refrained thus far from taking the drug and that he managed to survive the disease. His argument for switching to the drug rests on the observation that only 2 out of 1000 drug users died in the observational study, which he considers a rather small risk to take, given the effectiveness of the drug as a pain killer.

Conventional wisdom instructs us to warn Mr. B against consulting a nonexperimental study in matters of decisions, since such studies are marred with uncontrolled factors, which tend to bias effect estimates. Specifically, the death rate of 0.002 among drug users may be indicative of low tolerance to discomfort, or of membership in a medically-informed socio-economic group. Such factors do not apply to Mr. B, who did not use the drug in the past (be it by choice, instinct or ignorance), and who is now considering switching to the drug by rational deliberation. Conventional wisdom urges us to refer Mr. B to the randomized experimental study of Table 2, from which the death rate under controlled administration of the drug was evaluated to be $P(y_x) = 0.016$, eight times higher than 0.002.

What would his risk of death be, if Mr. B decides to start taking the drug? 0.2 percent or 1.6 percent?

The answer is that neither number is correct. Mr. B cannot be treated as a random patient in either study, because his history of not using the drug and his survival thus far puts him in a unique category of patients, for which the effect of the drug was not studied.[2] These two attributes provide extra evidence about Mr. B's sensitivity to the drug. This became clear already in Example 1, where we discovered definite relationships among these attributes – for some obscure reasons, terminal patients chose not to use the drug.

To properly account for this additional evidence, the risk should be measured through the counterfactual expression $PS = P(y_x|x', y')$; the probability that a patient who survived with no drug would have died had he/she taken the drug. The appropriate bound for this probability is given in Eq. (41):

$$0.002 \leq PS \leq 0.031$$

Thus, Mr. B's risk of death (upon switching to drug usage) can be as high as 3.1 percent; more than 15 times his intuitive estimate of 0.2 percent, and almost twice the naive estimate obtained from the experimental study.

However, if the drug can safely be assumed to have no death-preventing effects, then monotonicity applies, and the appropriate bound is given by Eq. (44), $PS = 0.002$, which coincides with Mr. B's intuition.

---

[2]The appropriate experimental design for measuring the risk of interest is to conduct a randomized clinical trial on patients in the category of Mr. B, that is, to subject a random sample of non-users to a period of drug treatment and measure their rate of survival.

## 6 Conclusion

This paper shows how useful information about probabilities of causation can be obtained from experimental and observational studies, with weak or no assumptions about the data-generating process. We have shown that, in general, bounds for the probabilities of causation can be obtained from combined experimental and nonexperimental data. These bounds were proven to be sharp and, therefore, they represent the ultimate information that can be extracted from statistical methods. We clarify the two basic assumptions – exogeneity and monotonicity – that must be ascertained before statistical measures such as excess-risk-ratio could represent attributional quantities such as probability of causation.

One application of this analysis lies in the automatic generation of verbal explanations, where the distinction between necessary and sufficient causes has important ramifications. As can be seen from the definitions and examples discussed in this paper, necessary causation is a concept tailored to a specific event under consideration (singular causation), whereas sufficient causation is based on the general tendency of certain event *types* to produce other event types. Adequate explanations should respect both aspects. Clearly, some balance must be made between the necessary and the sufficient components of causal explanation, and the present paper illuminates this balance by formally explicating the basic relationships between the two components. In Pearl (2000, chapter 10) it is further shown that PN and PS are too crude for capturing probabilities of causation in multi-stage scenarios, and that the structure of the intermediate process leading from cause to effect must enter the definitions of causation and explanation. Such considerations will be the subject of future investigation (See [Halpern and Pearl, 2000]).

Another important application of probabilities of causation is found in decision making problems. As was pointed out in Pearl (2000, pp. 217-219) and illustrated in Section 5, the counterfactual "$y$ would have been true if $x$ were true" can often be translated into a conditional action claim "given that currently $x$ and $y$ are false, $y$ will be true if we do $x$." The evaluation of such conditional predictions, and the probabilities of such predictions, are commonplace in decision making situations, where actions are brought into focus by certain eventualities that demand remedial correction. In troubleshooting, for example, we observe undesirable effects $Y = y$ that are potentially caused by other conditions $X = x$ and we wish to predict whether an action that brings about a change in $X$ would remedy the situation. The information provided by the evidence $y$ and $x$ is extremely valuable, and it must be processed before we can predict the effect of any action[3]. Thus, the expressions developed in this paper constitute bounds on the effectiveness of pending policies, when full knowledge of the state of affairs is not available, yet the pre-action states of the decision variable ($X$) and the outcome variable ($Y$) are known.

For these bounds to be valid in policy making, the data generating model must be time-invariant, that is, all probabilities associated with the model should represent epistemic uncertainty about static, albeit unknown boundary conditions $U = u$. The constancy of $U$ is well justified in the control and diagnosis of physical systems, where $U$ represents fixed, but unknown physical characteristics of devices or subsystems. The constancy approximation is also justified in the health sciences where patients' genetic attributes and physical characteristics can be assumed relatively constant between observation and treatment. For instance, if a patient in the example of Section 5 wishes to assess the risk of switching from no drug to drug, it is reasonable to assume that this patient's susceptibility to the drug remains constant through the interim period of analysis. Therefore, the risk associated with this patient's decision will be well represented by the counterfactual expression $PS = P(y_x | x', y')$, and should be assessed by the bounds in Eq. (41).

The constancy assumption is less justified in economic systems, where agents are bombarded by rapidly fluctuating streams of external forces ("shocks" in econometric terminology) and inter-agents stimuli. These forces and stimuli may vary substantially during the policy making interval and they require, therefore, detailed time-dependent analysis. The canonical violation of the constancy assumption occurs, of course, in quantum mechanical systems, where the indeterminism is "intrinsic" and memory-less, and where the existence of a deterministic relationship between the boundary conditions and measured quantities is no longer a good approximation. A method of incorporating such intrinsic indeterminism into counterfactual analysis is outlined in Pearl (2000, p. 220).

## Acknowledgements

We thank the referees for making useful suggestions on the first draft. Full version of this paper will appear in the Annals of Mathematics and AI, 2000. This research was supported in parts by grants from NSF, ONR and AFOSR and by a Microsoft Fellowship to the first author.

---

[3] Such processing have been applied indeed to the evaluation of economic policies [Balke and Pearl, 1995] and to repair-test strategies in troubleshooting [Breese and Heckerman, 1996]